\documentclass{article}

\usepackage{times}
\usepackage{graphicx} 
\usepackage{subfigure} 
\usepackage{multirow}
\usepackage{booktabs}

\usepackage{natbib}

\usepackage{algorithm}
\usepackage{algorithmic}
\renewcommand{\algorithmicrequire}{\textbf{Input:}}
\renewcommand{\algorithmicensure}{\textbf{Output:}}

\usepackage{hyperref}
\usepackage{url} 

\usepackage{amsmath,amsfonts,amssymb,amsthm}
\usepackage{bbm,bm,color}
\newtheorem{theorem}{Theorem}
\newtheorem{lemma}{Lemma}

\newtheorem{definition}{Definition}
\newtheorem{proposition}{Proposition}

\DeclareMathOperator*{\argmin}{arg\,min}

\newcommand{\rR}{\mathbb{R}} 
\newcommand{\numofproj}{P} 
\DeclareMathOperator{\diag}{diag}

\usepackage[accepted]{icml2015}

\icmltitlerunning{Learning Mixed Membership Mallows Models from Pairwise Comparisons}

\begin{document} 

\twocolumn[
\icmltitle{Learning Mixed Membership Mallows Models from pairwise comparisons}

\icmlauthor{Weicong Ding}{dingwc@bu.edu}
\icmlauthor{Prakash Ishwar}{pi@bu.edu}
\icmlauthor{Venkatesh Saligrama}{srv@bu.edu}
\icmladdress{Department of Electrical and Computer Engineering, Boston University, Boston MA 02215, USA}

\icmlkeywords{Approximate separability, nonnegative matrix factorization, Topic modeling, Mallows model, rank aggregation}

\vskip 0.3in
]

\begin{abstract} 
We propose a novel parameterized family of Mixed Membership Mallows Models (M4) to account for variability in pairwise comparisons generated by a heterogeneous population of noisy and inconsistent users. M4 models individual preferences as a user-specific probabilistic mixture of {\it shared} latent Mallows components. Our key algorithmic insight for estimation is to establish a statistical connection between M4 and topic models by viewing pairwise comparisons as words, and users as documents. This key insight leads us to explore Mallows components with a separable structure and leverage recent advances in separable topic discovery. 
While separability appears to be overly restrictive, we nevertheless show that it is an {\it inevitable} outcome of a relatively small number of latent Mallows components in a world of large number of items.
%
%
%
We then develop an algorithm based on robust extreme-point identification of convex polygons to learn the reference rankings, and is provably consistent with polynomial sample complexity guarantees. 
We demonstrate that our new model is empirically competitive with the current state-of-the-art approaches in predicting real-world preferences.
\end{abstract} 
\vspace*{-4ex}
\section{Introduction}
\label{sec:intro}
The problem of predicting preference for a diverse user-population arises in many applications including personal recommendation systems, e-commerce and information retrieval \citep{volkovs14a:ref, Lu11:ref, topicRank2:ref}. Pairwise comparisons of items by a heterogeneous and inconsistent population can now be observed and recorded over the web through transactions, clicks and check-ins for a large set of items. Our goal is to model, inference, and predict user behavior in pairwise comparisons.


%
This paper proposes a new {\it Mixed Membership Mallows Model} (M4) for pairwise comparisons that leverages the widely used mixture of Mallows model \citep[e.g.,][]{Lu11:ref, MxMallow14:ref}.
The building block of M4 is the popular Mallows distribution on permutations. The pmf of Mallows model is centered around a reference ranking and the deviation is captured by a dispersion constant \citep{mallows1957:ref}.
%
M4 naturally captures the {\it heterogeneous, inconsistent, and noisy} behavior by assuming each user's comparisons as a probabilistic mixture of a few {\it shared} latent Mallows components.
By design, the latent Mallows components capture the heterogeneous influencing factors in the population and the user-specific mixing weights reflect the influence of multiple latent factors on each user. 
Furthermore, the randomness of each Mallows component captures the fact that the same latent factor can consistently result in different outcomes on different users, more so far very similar items. 
Overall, M4 generalizes the clustering perspective in mixture of Mallows model into a {\it decomposition} modeling perspective that better fits the emerging web-scale observations.


%
%
%
%
%
%

The key contribution in this paper is to propose the  first {\it provable} and {\it polynomially efficient} approach for learning multiple Mallows components in mixed membership settings from pairwise comparisons. As a special case of M4, the mixture of Mallows model has received significant attention \citep{Lebanon02:ref, busse2007cluster, Lu11:ref, MxMallow14:ref}, yet theoretical guarantees are not clear except for special cases \cite{MxMallow14:ref}. We propose to learn M4 by reducing it to an instance of a probabilistic topic modeling \citep{Blei2012Review:ref}. Topic modeling for text corpus have been extensively studied but its connection to preference data is unclear. We view users as ``documents'', pairwise comparisons as ``words'', and the latent Mallows components as ``topics''. 
This leads us to the question of topic discovery viewed within the context of M4. 

The key technical contribution of our approach is to provably discover latent factors with a {\it non-exact separability} structure. Our approach is geometrically inspired by the recent work in exact separable topic discovery \citep[e.g.m][]{Arora2:ref, DDP:ref}, and we provably generalize it to approximately separability with finite degree of deviation. 
In M4, this requires for each Mallows component, there exist an item pair  such that item A is preferred over B with very high probability in that Mallows component and B is preferred over A with high probability under the other Mallows component. 
%
%
While it might appear restrictive, we show formally that approximate separability is inevitable and naturally arises from the fact that we have large set of items relative to the number of shared latent preferences. 
As a consequence, most large M4 are approximately separable. 
We then provably generalize the geometry property in solid angle from \cite{Ding14:ref} and establish guarantees for consistent estimation of reference rankings along with {\it polynomial} sample and computational complexity bounds. 
Our results only require the number of users to scale while allowing for the number of comparisons per user to be small.

\vspace*{-1ex}
\subsection{Related work}
\vspace*{-1ex}
\label{sec:related}
\begin{table*}[!htb]
\centering
\caption{Comparison to closely related works. ``vertices'' denote the prior that has non-zero probability only on the vertices of a simplex. }
\label{table:MixtureRankings}
\begin{tabular}{|c|c|c|c|c|c|}
\hline 
{\bf Method} & {\bf Observation} & {\bf Ranking component} & {\bf Prior }& {\bf Consistency} & {\bf Computation }\\ 
& {\bf type} & { (``Topic'')} & {\bf Distribution} &{\bf result} & {\bf complexity} \\
\hline
M4 & pairwise & Mallows  & general & provable & polynomial \\ 
\hline
\citet{topicRank2:ref}& pairwise & single ranking & general & provable & polynomial \\ 
\hline
\citet{gormley2008mixture:ref}& full & Plackett-Luce  & Dirichlet & not available & not available \\ 
\hline
\citet{Farias09:ref}& pairwise& single ranking & vertices & provable & combinatorial \\ 
\hline
\citet{Lu11:ref}& pairwise & Mallows & vertices & not available & not available \\
\hline
\citet{MxMallow14:ref}& top-3 rank & Mallows & vertices & provable & polynomial \\
\hline
\citet{MxMNL14:ref}& pairwise  & Bradely-Terry-Luce & vertices & provable & polynomial \\
\hline
\end{tabular} 
\end{table*}

Rank estimation from full or partial preferences has been extensively studied in different settings for decades \citep[][]{marden1996analyzing, Agarwal14:ref, volkovs14a:ref}. 
%
The family of mixture of ranking models have demonstrated superior modeling power to capture a heterogeneous population with noisy observations  \citep[e.g.,][]{Farias09:ref, MxMNL14:ref}. 
In these models, each user is associated with {\it one} ranking component sampled from a set of multiple ranking components hence the population can be clustered into heterogeneous preference types. 
%
The mixture of Mallows model has received significant attention \citep{Lebanon02:ref, busse2007cluster, Lu11:ref, MxMallow14:ref}. EM-based algorithms have been used for estimation from pairwise comparisons \citep{Lu11:ref} or full rankings \cite{busse2007cluster}.  Only recently, \citep{MxMallow14:ref} proposed a provably correct algorithm based on tensor decomposition that can handle a mixture of 2 Mallows model using the top-3 ranked items as the observations which, in effect, requires users to consider all items. This is impractical within the context of the target web-scale applications. 
Since the mixture of Mallows is special case of M4 by positing a specific prior on each user's mixing weights, our algorithm can thus be viewed as providing a powerful alternative approach for learning the mixture of Mallows model.
%
%
We note that mixture of Bradley-Terry-Luce (BTL) models \citep{MxMNL14:ref}, mixture of Plackett-Luce (PL) models \citep{MxRUT13:ref} have been studied.
%

%
Our model is closely related to \citep{topicRank:ref, topicRank2:ref} that validated the advantages of adopting the mixed membership perspective.
\citet{topicRank2:ref} models each latent ranking factor as a single permutation and is a {\it degenerate} special case of Mallows distribution over the permutations in M4. 
%
Therefore, while both \cite{topicRank2:ref} and M4 can capture the inconsistent behavior semming from the influence of multiple latent factors, M4 can further account for the inconsistency as the consequence of the randomness within each Mallows components. 
Our approach has similar polynomial time and sample guarantees as in \cite{topicRank2:ref}. 
We note that motivated by social choice application,  \citet{gormley2008mixture:ref} proposed another mixed membership ranking model where the latent ``topics'' are PL models. An MCMC based approach is used for estimation without theoretical guarantees.
Table.~\ref{table:MixtureRankings} summarizes all the closely related works. 
%
%
%

%
{\bf Connection to Separable Topic Discovery:}
A key motivation of our approach is the recent work on consistent and efficient topic discovery for topic matrices that have an {\it exact} separable structure \citep{Arora2:ref, Ding14:ref}. 
The exact separability has been exploited as a suitable approximation to many problems including topic modeling \cite{Arora2:ref} and ranking estimation \cite{topicRank2:ref}. 

Closely related to our technical settings is the so called near-separable structure where the observations are viewed as a noisy perturbation from some exact separable statistic. 
In the literature to-date, establishing provable guarantees requires the perturbation to go to zero via either data augmentation \citep{Arora2:ref, DDP:ref, topicRank2:ref} or improving Signal-to-Noise-Ratio \citep{Gillis2014:ref, Benson14:ref}.  
In contrast, the ideal statistic in our approach has a small but finite perturbation from the exactly separable ideal. Our provable guarantees require only a {\it finite} degree of approximate separability. We explicitly derive a sufficient condition that bounds on the degree of approximate separability.

%
\citet{bansal2014provable} recently proposed a provable approach that requires similar approximate separability as in our settings but requires a strong condition on the weight prior. In M4, it requires each user to have a dominant latent factor. In contrast, we only requires the second order moments of the prior to be full rank which is satisfied by many prior distributions \cite{Arora2:ref}. 

{\bf Rating based methods:}
Considerable work in preference prediction has focused on numerical ratings.
The most important idea is also to model the ratings as being influenced by a small number of latent factors shared by the population \citep[e.g.,][]{BPMF:ref}.
Although coming from a different feature space, our model shares the same mixed membership modeling perspective. 

%
The rest of the paper is organized as follows. Section~\ref{sec:generalmodel} introduces the M4 model. In Sec.~\ref{sec:topicGeometry}, we formally introduce the approximate separability and show that the set of approximate separable M4 models has an overwhelming probability. Section~\ref{sec:algorithm} summarizes the steps of our algorithm and the computational and sample complexity bounds. We demonstrate competitive performances on some semi-synthetic and real-world datasets in Sec.~\ref{sec:experiment}. 
%
\vspace*{-2ex}
\section{Mixed Membership Mallows Model}
\label{sec:generalmodel}
\begin{figure}[!htb]
\vspace*{-2ex}
\centering{
\includegraphics[width=0.8\linewidth]{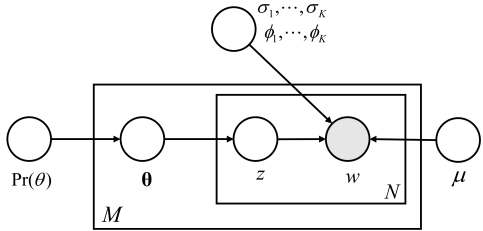}}
\vspace*{-4ex}
\caption{{\small Graphical representation of the proposed Mixed Membership Mallows Model. The boxes represent replicates. The bottom outer plate represents users, and the inner plate represents ranking tokens and comparisons of each user. }}
\label{fig:graphicalMallows}
\end{figure}
We now describe the generative process of the Mixed Membership Mallows Model (M4). To set up the problem,  we consider a universe of $Q$ items $\mathcal{U} = \{1,\ldots, Q\}$ and a population of $M$ users that each compares $N \geq 2$ pairs of items. We assume the item pairs to be compared, denoted by un-ordered pairs $\{i,j\}$, are drawn independently from some distribution $\mu$. The outcome of $n$-th pairwise comparison of user $m$ is denoted by an ordered pair $w_{m,n} = \left( i,j \right)$, if user $m$ compares item $i$ and $j$, and prefers $i$ over $j$. 

We first introduce the Mallows model \citep{mallows1957:ref}. In M4, let the $k$-th Mallows component define a probability distribution on the set of all permutations over the $Q$ items. It is parameterized by a reference ranking $\sigma_k$ and a dispersion parameter $\phi_k\in [0,1]$:
\vspace*{-1ex}
\begin{align}
p_{\text{M}}(\sigma \vert \sigma_{k},\phi_k)  =\phi_k^{d(\sigma,\sigma_k)} / Z_{k}
\end{align} 
\vglue -2ex
where $\sigma$ denotes an arbitrary permutation, $d(\sigma,\sigma_k)$ denotes the Kendall's tau distance between two permutations, and $Z_{k}$ is the normalization constant.
The generative process for the comparisons in M4 from user $m=1,\ldots, M$ is,
\vspace*{-2ex}
\begin{enumerate}
\item Sample ranking weight $\bm{\theta}_m \in \bigtriangleup^{K}$ from  prior $\Pr(\bm\theta)$.
\vspace*{-1ex}
\item For each comparison $n=1,\ldots, N$,
\vspace*{-1ex}
\begin{enumerate}
\item Sample a pair of items $\{i,j\}$ from $\mu$.
\item Sample a ranking token $z \in\{ 1,\ldots, K\} \sim\text{Multinomial}(\bm\theta_m)$
\item Sample a permutation $\sigma_{m,n}$ from $z$-th Mallows component with parameter $(\sigma_z,\phi_z)$
\item If $\sigma_{m,n}(i) < \sigma_{m,n}(j)$, then $w_{m,n} = \left( i, j \right)$, otherwise $w_{m,n} = \left( j, i \right)$
\footnote{ 
$\sigma(i)$ is the position of item $i$ in a ranking $\sigma$. Item $i$ is preferred over $j$ if $\sigma(i) < \sigma(j)$.
} 
\end{enumerate}
\vglue -2ex
\end{enumerate}
\vglue -2ex
Figure~\ref{fig:graphicalMallows} is the plate representation of M4. 
The mixing weights $\bm{\theta}_m$ over the  $K$ shared Mallows components characterize each user.
We denote by $W\times M$ matrix $\mathbf{X}$ for the empirical observations. Its $W=Q(Q-1)$ rows are indexed by all the ordered pairs $(i,j)$. $X_{(i,j),m}$ denotes the number of times that user $m$ prefers item $i$ over $j$. 
Given $\mathbf{X}$ and $K$, the primary problem in this paper is to learn the parameters of the shared latent Mallows component.

{\bf Reduction to Topic Modeling} 

We show that the problem of learning model parameters in M4 can be formally reduced to topic discovery in an equivalent topic model. 
To establish the connection, we first consider the distribution on the pairwise comparisons $w_{m,n}$,
%
\vspace*{-1ex}
\begin{align*}
 p(w_{m,n} = (i,j) \vert \bm{\theta}_m) =  \mu_{i,j} \sum_{k=1}^{K} \sum_{~\sigma(i)<\sigma(j)} p_{\text{M}}(\sigma \vert \sigma_{k}, \phi_k) \theta_{k,m}
\end{align*}
\vglue -2ex
where $\mu_{i,j} = \mu_{j,i} >0$ is the probability of comparing item $i$ and $j$. For further reference, we define  {\bf ranking matrix} to be a $W\times K$ dimension matrix $\bm{\beta}$ whose entries are,
\vspace*{-1ex}
\begin{align}
\label{eqa:rankingmatrix}
\beta_{(i,j),k} :=  \sum_{\sigma:~\sigma(i)<\sigma(j)} p_{\text{M}}(\sigma \vert \sigma_{k}, \phi_k)
\end{align}
\vglue -2ex
Statistically, $\beta_{(i,j),k}$ represents the probability that item $i$ is preferred over item $j$ if the ranking is sampled from the $k$-th Mallows component. 
The $k$-th column of $\bm{\beta}$ therefore captures the pairwise comparison behavior induced by the $k$-th Mallows distribution and is a function determined only by $\sigma_k, \phi_k$. For convenience, we also  define a $W\times K$ matrix $\mathbf{B}$ as $B_{(i,j),k} =\mu_{i,j} \beta_{(i,j),k}$.
Therefore, the conditional probability of the comparisons can be simplified as,
\vspace*{-2ex}
\begin{align}
\label{eqa:prob_genearl_ranking}
p\left( w_{m,n} = (i,j)\vert \bm{\theta}_m \right) = \sum_{k=1}^{K} B_{(i,j),k}\theta_{k,m}
\end{align}
\vglue -2ex

Before we connect to topic modeling, we summarize the properties of the ranking matrix $\bm{\beta}$ that enable us to infer the Mallows parameters directly from $\mathbf{B}$: 
\begin{proposition}
\label{prop:beta_properties}
Let the ranking matrix $\bm{\beta}$ be defined as in Eq.~\eqref{eqa:rankingmatrix}, and $\sigma_{k},\phi_k$'s are parameters of the $K$ Mallows distribution. Then, $\forall (i,j)$ and $\forall k$, we have,
\vspace*{-2ex}
\begin{enumerate}
%
\item[a.] $\beta_{(i,j),k} = \frac{B_{(i,j),k}}{B_{(i,j),k} + B_{(j,i),k}}$
\vspace*{-1ex}
\item[b.] If $\sigma_k(i) < \sigma_k(j)$ and $\phi_k < 1$, $\beta_{(i,j),k} > 0.5 > \beta_{(j,i),k}$
\item[c.] If $\sigma_k(j) = \sigma_k(i) + 1$ and $\phi_k < 1$, $1/\beta_{(i,j),k}  = 1+\phi_k$
\end{enumerate}
\end{proposition} 
\vspace*{-1ex}
First, by Prop.~\ref{prop:beta_properties}~a., we can directly infer $\bm{\beta}$ from $\mathbf{B}$. Second, by Prop.~\ref{prop:beta_properties}~b., one can infer the relative position of any two items in the reference rankings $\sigma_1,\ldots,\sigma_K$ by comparing the entries in $\bm{\beta}$ with $1/2$. 
Therefore, if the estimation error in $\bm{\beta}$ is element-wise small and $\phi_k <1$, then, all the pairwise relations in the $K$ reference rankings can be correctly inferred hence the total rankings. 
Furthermore, the dispersion can be estimated using Prop.~\ref{prop:beta_properties}~c.
\footnote{If $\phi_k =1$, the $k$-th Mallows component is the uniform distribution  and is un-identifiable. We consider $\phi_k<1$ in this paper.}.
In sum, we can learn all the model parameters from $\mathbf{B}$. For the rest of this paper, we focus on learning $\mathbf{B}$.

We note that Eq.~\eqref{eqa:prob_genearl_ranking} shares the same structure as in probabilistic topic modeling \cite{Blei2012Review:ref, MMLVM14:ref}. We consider a topic model on a set of  $M$ documents, each composed of $N\geq 2$ words that are drawn from a vocabulary of size $W$,  with a $W\times K$ dimension topic matrix $\bm{\beta}^{\text{TM}}$, and the document-specific topic weights $\bm{\theta}_{m}^{\text{TM}}$ sampled independently from a topic prior $\Pr^{\text{TM}}(\theta)$.
The conditional distribution on $w_{m,n}^{\text{TM}}$, the $n$-th word in document $m$, is
\vspace*{-2ex}
\begin{align}
\label{eqa:prob_topicmodel}
p(w_{m,n}^{\text{TM}} = i  \vert \bm{\theta}^{\text{TM}}_m) & = \sum_{k=1}^{K} \beta^{\text{TM}}_{i,k}\theta^{\text{TM}}_{k,m}
\end{align}
\vglue -2ex
where $i=1,\ldots, W$ are distinct words in the vocabulary. Noting that $\mathbf{B}$ is also column-stochastic, we have, 
\begin{lemma}
\label{lem:statequivalent}
The proposed Mixed Membership Mallows Model is statistically equivalent to a topic model whose topic matrix $\bm{\beta}$ is set to be $
\mathbf{B}$ and the topic prior to be $\Pr(\theta)$.
\end{lemma}
\vspace*{-2ex}
\begin{proof}
We consider the distribution on the observations in both model, i.e, the distribution on the outcomes of pairwise comparisons $\mathbf{w} =\{w_{m,n}\}$ in M4 and the words $\mathbf{w}^{\text{TM}} =\{w_{m,n}^{\text{TM}}\}$ in topic model. Note that each user is independent conditioned on $\bm{\theta}_m$, from the conditional probabilities in Eq.~\eqref{eqa:prob_genearl_ranking} and \eqref{eqa:prob_topicmodel}, we have,
\vspace*{-1ex}
\begin{align*}
p(\mathbf{w} \vert \mathbf{B} ) & = \prod_{m=1}^{M} \int
p(w_{m,1},\ldots, w_{m,N}\vert \bm{\theta}_m, \mathbf{B})
\Pr(\bm{\theta}_m) d \bm{\theta}_m \\
& = \prod_{m=1}^{M} \int \left( \prod_{n=1}^{N}
\sum_{k=1}^{K}B_{w_{m,n},k}\theta_{k,m} \right) \Pr(\bm{\theta}_m) d
\bm{\theta}_m \\
%
%
&= p(\mathbf{w}^{\text{TM}} \vert \bm{\beta}).
\end{align*}
\vglue -2ex
which is the same as in topic models \cite{Blei2012Review:ref}.
\end{proof}
\vspace*{-2ex}
Thus, the estimation problem in M4 can be solved by first learning $\mathbf{B}$ using any topic modeling algorithms, and then estimating the parameters of the shared Mallows components using Prop.~\ref{prop:beta_properties}. Before we discuss our approach in detail in next section, we consider the relation between M4 and other ranking models. We highlight that the proposed M4 is a much more general family that subsumes a few existing ranking models as special cases: 
\begin{proposition}
\label{prop:relation2otherModels}
In Mixed Membership Mallows Model,
\vspace*{-2ex} 
\begin{enumerate}
\item If the dispersion parameters $\phi_k \rightarrow 0$, then, each Mallows component has non-zero probability only on the reference ranking $\sigma_k$, and the Mixed Membership Mallows Model reduces to topic modeling framework proposed in \cite{topicRank2:ref}. 
\vspace*{-1ex}
\item If the topic prior $\Pr(\theta)$ has non-zero probability only on the vertices of $K$-dimension simplex, then, each user can only be influenced by one Mallows components and the Mixed Membership Mallows Model reduces to the mixture of Mallows model \cite{Lu11:ref, MxMallow14:ref}
\end{enumerate}
\end{proposition}
\vspace*{-2ex}
\section{A Geometric Approach}
\label{sec:topicGeometry} 
We discuss in this section the key geometric insights of our approach. We leverage the recent works in separable topic discovery that come with consistency and efficiency guarantees \citep[][etc.]{Arora2:ref, Ding14:ref, Kumar13:ref,bansal2014provable}. 
The consistency is favorable here since we are not enforcing the estimation to be valid total rankings. 
To be precise, we exploit the geometric property of the second-order moments of the columns of $\mathbf{X}$, i.e., a co-occurrence matrix of pairwise comparisons, which can be estimated consistently:
\begin{lemma}
\label{lem:2ndOrder1}
If $\widetilde{\mathbf{X}}$ and $\widetilde{\mathbf{X}}^{\prime}$ are
obtained from $\mathbf{X}$ by first splitting each user's  comparisons into two independent halves and then re-scaling the rows to make them
row-stochastic, then
\vspace*{-1ex}
\begin{equation}
M \widetilde{\mathbf{X}}^{\prime} \widetilde{\mathbf{X}}^{\top}
\xrightarrow[\mbox{almost surely}]{M \rightarrow\infty}
\bar{\bm{B}}\bar{\mathbf{R}} \bar{\bm{B}}^{\top} =: \mathbf{E},
\vspace*{-1ex}
\end{equation}
where 
$\bar{\bm{B}} = \diag^{-1}(\bm{B}\mathbf{a})\bm{B}\diag(\mathbf{a})$,
$\bar{\mathbf{R}} = \diag^{-1}(\mathbf{a})
\mathbf{R}\diag^{-1}(\mathbf{a})$, and
$\mathbf{a}$ and $\mathbf{R}$ are, respectively, the $K\times 1$
expectation and $K\times K$ correlation matrix of the weight vector
$\bm{\theta}_m$.
\end{lemma} 
In this paper, we always assume that $\mathbf{R}$ (the $K\times K$ topic co-occurrence matrix) has full rank which is satisfied by many important prior distributions \cite{Arora2:ref}. 
\vspace*{-1ex}
\subsection{Approximate Separability}
\vspace*{-1ex}
\label{subsec:near_separable}
The consistent separable topic discovery approaches \citep[e.g.,][]{Arora2:ref, Ding14:ref} require the ranking matrix $\bm{\beta}$ to be {\it exactly separability}, i.e., for each $k$, there exist some {\it novel} rows (i.e., ordered pairs $(i,j)$) such that $\beta_{(i,j),k} > 0$ and $\beta_{(i,j),l}=0, \forall~l\neq k$. If this exact separability condition holds, the row vectors in $\mathbf{E}$ of the novel pairs will be extreme points of the convex hull formed by all row vector of $\mathbf{E}$ (the shaded dash circles in Fig.~\ref{fig:extreme}).  

By the definition of the ranking matrix in Eq.~\eqref{eqa:rankingmatrix}, for $\phi_k>0$, none of the entries in the ranking matrix $\bm{\beta}$ is identically zero. Hence exact separability can not be satisfied. However, recall that $\beta_{(i,j),k}$ is the probability of preferring item $i$ over $j$ in the $k$-th Mallows component, by the property of the Mallows distribution, $\beta_{(i,j),k}$ will be very close to 0 if the position of item $j$ in the reference ranking $\sigma_k$ is higher than $i$ by a large margin. Explicitly, 
\begin{proposition}
\label{prop:beta_close_to_zero}
Let $\sigma_k(i)$ and $\sigma_k(j)$ be the positions of items $i$ and $j$ in the reference ranking $\sigma_k$ of the $k$-th Mallows component and $\phi_k < 1$. If $\sigma_k(i) > \sigma_k(j)$ and $L = \sigma_k(i)-\sigma_k(j) + 1$, then,
\vspace*{-1ex}
\begin{equation}
\beta_{(i,j),k} \leq \frac{L\phi_k^{L-1}}{1+L\phi_k^{L-1}}
\end{equation}
\end{proposition}
\vglue -1ex
Since $\phi_k < 1$, if $L$ increases, the corresponding $\beta_{(i,j),k}$ is arbitrarily close to 0. Motivated by this observation in Prop.~\ref{prop:beta_close_to_zero}, we propose to consider the ranking matrix $\bm{\beta}$ that is {\it approximately separable}:
\begin{definition}
\label{def:approx_separable_condition}
{\bf ($\lambda$-Approximate Separability)} A $W\times K$ non-negative matrix ${\bm \beta}$ is $\lambda$-approximately separable for some constant $\lambda \in [0,1)$, if $\forall k=1,\ldots, K$, there exists at least one row (i.e., ordered pair) $(i,j)$ such that $\beta_{(i,j),k} > 0$ and $\beta_{(i,j),l} \leq \lambda \beta_{(i,j),k}$, $\forall l\neq k$. 
\end{definition}
The $\lambda$-approximate separability requires the existence of ordered pairs that having negligible probability in all-but-one Mallows components, i.e., the row weights concentrates predominantly in one column (see Fig.~\ref{fig:extreme}). 
We will refer to such pairs (rows of $\bm{\beta}$) as $\lambda$-approximate novel pairs (rows) for each latent factor.
By Prop.~\ref{prop:beta_close_to_zero} for M4, the approximate separability boils down to the existence of  pairs of items $\{i,j\}$ such that $i$ is uniquely preferred over $j$ in one reference ranking, while $j$ is ranked higher than $i$ {\it by a large margin} in all other reference rankings.

For small $\lambda$, this seems to be a very restrictive condition on the shared latent Mallows distribution. However, as we show shortly in the next section, most M4 models are approximately separable for small constant $\lambda >0$ if the number of items $Q$ scales sufficiently faster than $K$.
Therefore, only a negligible fraction of models in M4 do not satisfy approximate separability.
\vspace*{-1ex}
\subsection{Inevitability of the Approximate Separability}
\vspace*{-1ex}
We investigate the probability that approximate separability is satisfied when we draw uniformly from M4. 
Specifically, we sample the $K$ reference rankings $\sigma_{k}$ uniformly i.i.d from the set of all permutations, and set $\phi_k \leq \phi <1, \forall k$. We have,
\begin{lemma}
\label{lem:separable_withhighprob}
Let the $K$ reference rankings $\sigma_1,\ldots, \sigma_K$ be sampled i.i.d uniformly from the set of all permutations, and the dispersion parameters $\phi_k <\phi < 1, k=1,\ldots, K$.
Then, the probability that the ranking matrix $\bm{\beta}$ being $\lambda$-approximately separable is at least 
\vspace*{-1ex}
\begin{align}
\label{eqa:p1inTMM}
1 - K \exp( - \frac{Q}{ L(\phi,\lambda)^{2K -1}} )
\end{align} 
\vglue -2ex
where $L(\phi,\lambda) = \text{ceil}\left( (1 + \frac{\log(\lambda)}{\log(\phi)})(1+\epsilon) \right)$ for some positive constant $\epsilon$, and $\text{ceil}(x)$ is the minimum integer that is no smaller than $x$.  
\end{lemma}
Therefore, for $Q\gg K$, the ranking matrix $\bm{\beta}$ is going to be approximately separable with high probability.
$L$ is determined by $\log(\lambda)/\log(\phi)$, and would be small for very small $\lambda$ because of the logarithmic dependence. The proof exploits the property illustrated in Prop.~\ref{prop:beta_close_to_zero} and is deferred to the supplementary section. We note that the result in Eq.~\eqref{eqa:p1inTMM} is only a loose upper bound on non-separable probability. 

We point out that by definition, approximate separability of $\bm{\beta}$ is equivalent to $\mathbf{B}$. Therefore $\mathbf{B}$ is also approximately separable with high probability. 
\vspace*{-1ex}
\subsection{Robust Novel Pair Detection}
\label{sec:robust_geomertry}
\begin{figure}[!htb]
\vglue -1ex
\centering{
\includegraphics[width=0.9\linewidth]{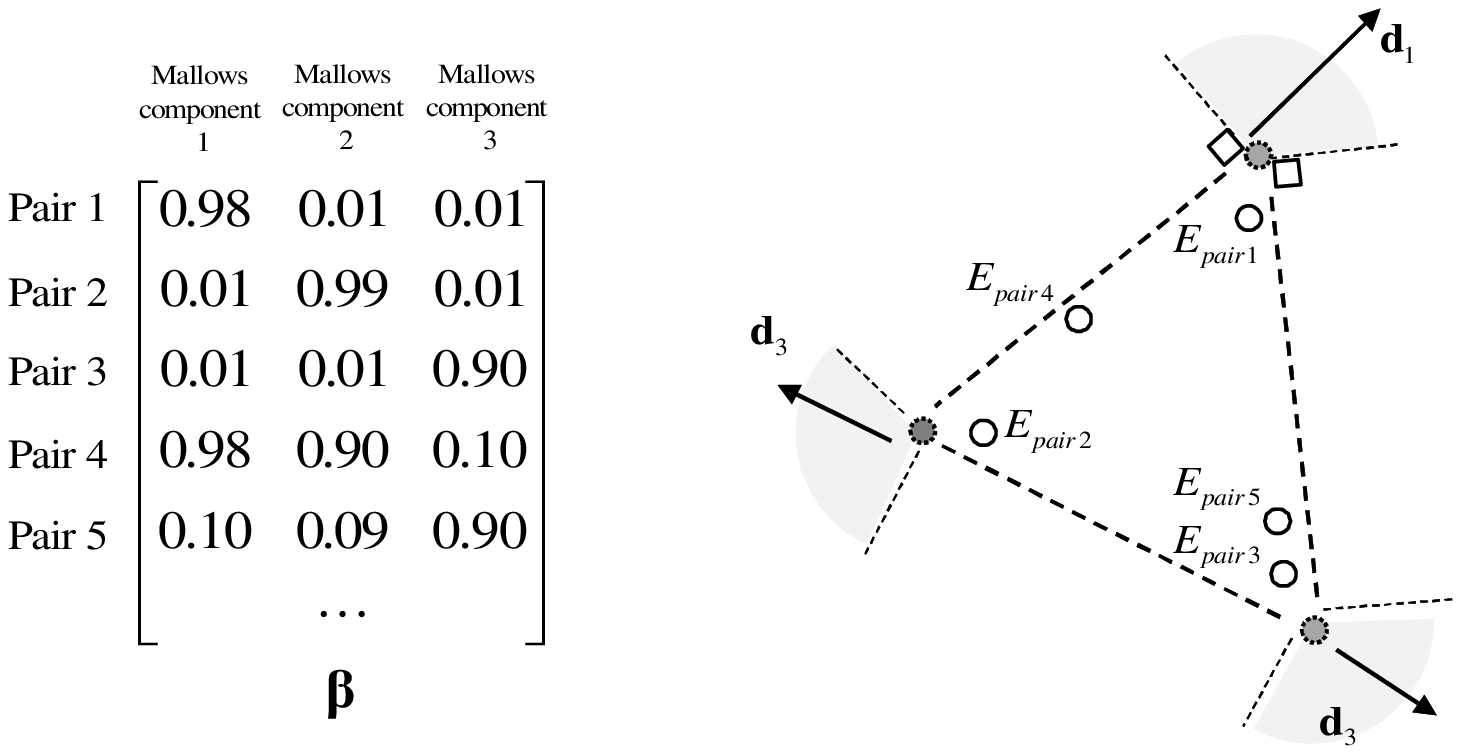}}
\vglue -1ex
\caption{\small{ An example of approximate separable $\bm{\beta}$ with $K=3$, and the underlying geometry of the row vectors of $\mathbf{E}$. Pair 1, 2, 3 are approximate novel pairs for Mallows component 1, 2, and 3. The shaded dash circles represent the ideal extreme points with exact separable $\bm{\beta}$ and the shaded regions depict their solid angles. The numbers in $\bm{\beta}$ are from $\phi_k=0.1$. $\beta_{(i,j),k}\approx 0.01$ when $L=3$, $\beta_{(i,j),k}\approx 0.1$ when $L=2$. $L = \sigma_k(i) - \sigma_k(j) + 1$. }}
\label{fig:extreme}
\end{figure}
\vspace*{-1ex}
Recall that when $\bm{\beta}$ is exactly separable, the novel rows in $\mathbf{E}$ are extreme points (shaded dash circles in Fig.~\ref{fig:extreme}). 
If $\bm{\beta}$ is $\lambda$-approximate separable with small enough $\lambda$, the rows $\mathbf{E}$ can be viewed as a small perturbation from the ideal case. As a result, the rows corresponding to the approximate novel pairs will be inside the ideal convex hull and are close to the ideal extreme points ($\mathbf{E}_{\text{pair} 1}$, $\mathbf{E}_{\text{pair} 2}$, and $\mathbf{E}_{\text{pair} 3}$ in Fig.~\ref{fig:extreme}). On the other hand, the non-novel rows could become extreme points but would be close to the convex hull formed by the approximate novel rows (e.g., $\mathbf{E}_{\text{pair} 4}$ in Fig.~\ref{fig:extreme}). 

We detect the approximate novel pairs as the most ``extreme'' rows of $\mathbf{E}$ based on a robust geometric measure, the normalized {\bf Solid Angle} subtended by extreme points (see Fig.~\ref{fig:extreme}) \cite{Ding14:ref}. 
%
Statistically, it is the probability that a row vector $\mathbf{E}_{(i,j)}$ has the maximum projection value along an isotropically distributed direction $\mathbf{d}\in\mathbb{R}^{W\times 1}$:
\begin{align}
\label{eqa:solidangle}
\nonumber q_{(i,j)} \triangleq p \{ & \forall (s,t):
\Vert \mathbf{E}_{(i,j)} - \mathbf{E}_{(s,t)}\Vert \geq \zeta, \\
&\qquad\qquad \mathbf{E}_{(i,j)} \mathbf{d} > 
\mathbf{E}_{(s,t)} \mathbf{d}  \}
\end{align}
When $\bm\beta$ is exact separable,  $q_{(i,j)}=0$ for non-novel pairs and are strictly positive for novel pairs. 
When the deviation introduced by $\lambda$-approximate separability is small, the solid angle for approximate novel pairs will be close to that of the ideal extreme points. 
For the non-novel pairs that become extreme points due to $\lambda$-approximate separability ($\mathbf{E}_{\text{pair} 4}$ in Fig.~\ref{fig:extreme}), the associated solid angles will be close to 0 since that it is very close to the convex hull formed by the rows of approximate separable pairs. 
In summary, if we sort the solid angles for all rows in $\mathbf{E}$,  the ones with largest solid angles must corresponds to $c\lambda$-approximate novel pairs for some constant $c$ and properly defined $\zeta$ in Eq.~\eqref{eqa:solidangle}. 

By definition in Eq.~\eqref{eqa:solidangle}, the solid angles can be consistently approximated using a few i.i.d isotropic  $\mathbf{d}$'s and an asymptotically consistent estimate of $\mathbf{E}$ \cite{Ding14:ref}. 
Once all the approximate novel pairs for $K$ distinct Mallows components are identified, $\mathbf{B}$ and therefore the model parameters can be estimated using constrained linear regression \cite{Arora2:ref, Ding14:ref} and Prop.~\ref{prop:beta_properties}. 
Given the estimated parameters of the ranking matrix $\bm{\beta}$, we can infer the user-specific preference weight $\bm{\theta}_m$ and evaluate the prediction probability of new comparisons using standard inference in topic modeling \cite{Blei2012Review:ref}.  
\vspace*{-2ex}
\section{Algorithm and Complexity Bounds}
\label{sec:algorithm}
The main steps of our approach are outlined in Alg.~\ref{alg:highlevel} and expanded in detail in Alg.~\ref{alg:rp},~\ref{alg:esttopic} and ~\ref{alg:post}.
Alg.~\ref{alg:rp} detects all the approximate novel pairs for the $K$ distinct latent components. Alg.~\ref{alg:esttopic} estimates matrix $\mathbf{B}$ using constrained linear regression followed by row scaling.
Alg.~\ref{alg:post} further infers the model parameters from $\widehat{\mathbf{B}}$ using Prop.~\ref{prop:beta_properties}. In particular, Step 2 of Alg.~\ref{alg:post} estimates all the pairwise relations in the reference rankings where $\sigma_{(i,j),k} =\mathbb{I}(\sigma_{k}(i) < \sigma_{k}(j))$ (which is an equivalent representation of a total ranking), and Step 4 estimates $\phi_k$. 
%
\begin{algorithm}[!htb]
\caption{M4 Estimation (Main Steps)}
\label{alg:highlevel}
\begin{algorithmic}[1]
\REQUIRE Pairwise comparisons $\widetilde{\mathbf X}$,
$\widetilde{\mathbf X}^{\prime} (W\times M)$ (defined in Lemma~\ref{lem:2ndOrder1}); Number of latent components $K$;
Number of projections $P$; Tolerance parameters $\zeta,\epsilon >0$
\ENSURE Reference ranking $\widehat{\sigma}_k$ and dispersion $\widehat{\phi}_k$, $k=1,\ldots,K$
\STATE Novel Pairs
$\mathcal{I}\leftarrow$NovelPairDetect($\widetilde{\mathbf
  X},\widetilde{\mathbf X}^{\prime}, K, \numofproj, \zeta$)
\STATE $\widehat{\mathbf{B}} \leftarrow$EstimateRankingMatrix($\mathcal{I},
\mathbf{X}, \epsilon$)
\STATE $\widehat{\sigma}_1,\ldots, \widehat{\sigma}_K,\widehat{\phi}_1,\ldots, \widehat{\phi}_K \leftarrow$PostProcess($\widehat{\mathbf{B}}$)
\end{algorithmic}
\end{algorithm}
%
%
\vspace*{-2ex}
\begin{algorithm}[!htb]
\caption{NovelPairDetect(via Random Projections)}
\label{alg:rp}
\begin{algorithmic}
\REQUIRE $\widetilde{\mathbf X}$, $\widetilde{\mathbf X}^{\prime}$;
number of rankings $K$; number of projections $P$; tolerance $\zeta$;
\ENSURE $\mathcal{I}$: The set of all novel pairs of $K$ distinct rankings.
\STATE $\widehat{\mathbf{E}} \leftarrow M \widetilde{\mathbf
  X}^{\prime}\widetilde{\mathbf X}^{\top}$
\STATE $\forall (i,j)$, $\mathcal{J}_{(i,j)} \leftarrow \lbrace (s,t):
\Vert \widehat{E}_{(i,j)}
-2\widehat{E}_{(s,t)}\Vert  \geq \zeta/2
\rbrace$,
\FOR {$r=1,\ldots,P$}
\STATE Sample ${\mathbf d}_r \in\rR^{W}$ from an isotropic prior
	\STATE $\hat{q}_{(i,j),r} \leftarrow \mathbb{I}\lbrace \forall
        (s,t)\in\mathcal{J}_{(i,j)}, ~ \widehat{\mathbf{E}}_{(s,t)}
               {\mathbf d}_r \leq \widehat{\mathbf{E}}_{(i,j)}
               {\mathbf d}_r \rbrace$ , $\forall (i,j)$
\ENDFOR
\STATE $\hat{q}_{(i,j)} \leftarrow \frac{1}{P} \sum_{r = 1}^{{P}}
\hat{q}_{(i,j), r}$, $\forall (i,j)$
\STATE $k \leftarrow 0$,$l \leftarrow 1$, and $\mathcal{I}
\leftarrow\emptyset$
\WHILE {$k \leq K$}
\STATE $(s,t) \leftarrow$ index of the $l^{\text{th}}$ largest
value among $\hat{q}_{(i,j)}$'s
\IF {$(s,t) \in \bigcap_{(i,j)\in\mathcal{I}} \mathcal{J}_{(i,j)}$}
\STATE {$\mathcal{I} \leftarrow \mathcal{I} \cup \{(s,t)\}$, $~~k
\leftarrow k + 1$}
\ENDIF
\STATE $l \leftarrow l + 1$
\ENDWHILE
\end{algorithmic}
\end{algorithm}
%
%
\begin{algorithm}[!htb]
\caption{Estimate Ranking matrix}
\label{alg:esttopic}
\begin{algorithmic}
\REQUIRE $\mathcal{I} = \{(i_1,j_1),\ldots, (i_K, j_K)\}$ the set of
novel pairs of $K$ rankings; ${\mathbf X}$, ${\mathbf X}^{\prime}$;
precision $\epsilon$
\ENSURE $\widehat{{\bm{B}}}$ as the estimate of ${\bm{B}}$. 
\STATE 
$
{\mathbf Y} = (\widetilde{\mathbf X}_{(i_1,j_1)}^{\top}, \ldots,
\widetilde{\mathbf X}_{(i_K,j_K)}^{\top})^{\top},$
\STATE
$
{\mathbf Y^{\prime}} = (\widetilde{\mathbf
  X}_{(i_1,j_1)}^{{\prime}\top}, \ldots, \widetilde{\mathbf
  X}_{(i_K,j_K)}^{{\prime}\top})^{\top}
$
\FORALL {$(i,j)$ pairs}
\STATE Solve $\widehat{\bm{\beta}}_{(i,j)} \leftarrow \argmin\limits_{\mathbf{b}} M
       (\widetilde{\mathbf X}_{(i,j)} - {\mathbf b} {\mathbf Y})
       (\widetilde{\mathbf X}^{\prime}_{(i,j)} - {\mathbf b} {\mathbf
         Y}^{\prime})^{\top} $
\STATE Subject to $b_k \geq 0, \sum_{k=1}^{K} b_k = 1$, With precision
$\epsilon$
\STATE $\widehat{\bm{\beta}}_{(i,j)} \leftarrow (\frac{1}{M} {\mathbf X}_{(i,j)} {\mathbf 1}) \widehat{\bm{\beta}}_{(i,j)}$
\ENDFOR
\STATE $\widehat{\bm{B}} \leftarrow$column normalize $\widehat{\bm{    \beta}}$
\end{algorithmic}
\end{algorithm}
%
\vspace*{-2ex}
\begin{algorithm}[!htb]
\caption{Post Processing}
\label{alg:post}
\begin{algorithmic}[1]
\REQUIRE $\widehat{\mathbf{B}}$ as the estimate of $\mathbf{B}$
\ENSURE $\widehat{\sigma}_{k},\widehat{\phi}_{k}, k =1,\ldots, K$ 
\STATE $\widehat{{\beta}}_{(i,j),k}\leftarrow
\frac{\widehat{{B}}_{(i,j),k}}{\widehat{{B}}_{(i,j),k} +
  \widehat{{B}}_{(j,i),k}}$, $\forall i,j\in\mathcal{U}, \forall k$
\STATE $\widehat{{\sigma}}_{(i,j),k} \leftarrow
\text{Round}[\widehat{{\beta}}_{(i,j),k}]$, $\forall
i,j\in\mathcal{U}, \forall k$
\STATE $\widehat{\sigma}_k \leftarrow \text{GlobalRank}(\widehat{\sigma}_{(i,j),k},\forall i,j)$~$\forall k$ ({\small First count the number of times each item wins in all pairwise comparison and then sort.})
\STATE $\widehat{\phi}_k \leftarrow \frac{1}{Q-1} \sum_{i=1}^{Q-1} \frac{1}{\widehat{\beta}_{(\sigma_k^{-1}(i), \sigma_k^{-1}(i+1)),k}} -1$,~$\forall k$ ({\small $\sigma_k^{-1}(i)$ is the item in the $i$-th position in ranking $\sigma_k$.})
\end{algorithmic}
\end{algorithm}

Our approach has an overall polynomial computation complexity in all model parameters,
\begin{theorem}
\label{thm:computation}
The running time of Algorithm~\ref{alg:highlevel} is $\mathcal{O}(MNK
+ Q^{2}K^{3})$.
\end{theorem}
\vspace*{-1ex}
The proofs are in supplementary. We note that the term $Q^{2}K^{3}$ is a loose upper bound for linear regression in Alg.~\ref{alg:esttopic}.
We also derive the sample complexity bounds for Alg.~\ref{alg:highlevel} which
is also polynomial in all model parameters and $\log(1/\delta)$ where
$\delta$ is the error probability.
Formally, 
\begin{theorem}
\label{thm:sample}
Let the ranking matrix $\bm{\beta}$ be $\lambda$-approximate separable and the second order moments $\mathbf{R}$ of ranking prior to be full rank. If
\vspace*{-2ex}
\begin{align}
\label{eqa:lambda_min}
\lambda \leq  \frac{a_{\min} \kappa (1-\phi) q_{\wedge}}{8 K^{2} a_0 \sqrt{\log(W/q_{\wedge})}}
\end{align}
\vglue -2ex
and $M,P \rightarrow \infty$, then, Algorithm~\ref{alg:highlevel} can consistently recover all the reference rankings of the latent Mallows distributions. Moreover, $\forall \delta>0$, if 
\vspace*{-1ex}
\begin{align*}
M \geq \max \Biggl\{\frac{640 W^2 \log(3W/\delta)}{ N \eta^4 d^2 q_{\wedge}^{2}}, ~
\frac{320 W \log(3W/\delta)}{N \eta^4 \lambda_{\min}^{2} a_{\min}^{2} (1-\phi)^{2}} \Biggr\}
\end{align*}
\vglue -1ex
and for 
\begin{equation*}
P \geq 32 \frac{\log(3W/\delta)}{q_{\wedge}^2}
\end{equation*}
the proposed algorithm fails with probability at most $\delta$. 
The other model parameters are defined as follows: $\eta = \min_{1\leq w\leq W} [\mathbf{B}\mathbf{a}]_w$; $a_{\max}$, $a_{\min}$ are the max/min of entries of $\mathbf{a}$; $a_0 =\max_{i,j} a_{i}/a_{j}$; $\mathbf{Y} = \bar{\mathbf{R}}\bar{\mathbf{B}}$; $\kappa = \lambda_{\min} / \lambda_{\max}$ is the condition number of $\bar{\mathbf{R}}$; $q_{\wedge}$ be the minimum normalized solid angle formed by row vectors of $\mathbf{Y}$; $d = 6\kappa / K$; $\phi_k \leq \phi <1$. $N$ is the number of comparisons of each user.
\end{theorem}
\vspace*{-1ex}
The detailed proofs are summarized in the supplementary file. Eq.~\eqref{eqa:lambda_min} provides an explicit sufficient upper bound on the required $\lambda$-approximate separable degree. It is roughly inverse polynomial in $K$. By Prop.~\ref{prop:beta_close_to_zero}, the margin $L$ required to satisfy $\lambda$ in Eq.~\eqref{eqa:lambda_min} should scale as $O(\log(K))$ which is small. 

We note that in the complexity bounds, the term $1-\phi$ represents the spread of the Mallows components and determines  the hardness of estimation: for smaller $\phi$, $\lambda$ can be larger and the required $M$ is smaller. When $\phi\rightarrow 1$, Eq.~\eqref{eqa:lambda_min} reduces to $\lambda = 0$ and $M\geq \infty$ which is not achievable and the corresponding Mallows distribution is un-identifiable. 
\vspace*{-2ex}
\section{Experimental validation}
\label{sec:experiment}
We conduct experiments to validate the performance of our proposed approach when the M4 assumptions are satisfied on semi-synthetic dataset, and then demonstrate that the proposed M4 can indeed effectively capture the preference behavior in real-world datasets. 
In all experiments, we used the suggested settings by \cite{Ding14:ref}. Specifically, the number of random projections $P = 150\times K$, the tolerance $\zeta=0.05$ in Alg.~\ref{alg:rp} and  $\epsilon = 10^{-4}$ in Alg.~\ref{alg:esttopic}.
\vspace*{-1ex}
\subsection{Semi-synthetic Simulation}
\vspace*{-1ex}
We generate synthetic examples according to proposed M4 and evaluate the proposed algorithm using reconstruction error measured by the {\bf Kendall's tau distance} between the estimated reference rankings and the ground-truth. Since our estimation is up to a column permutation, we align the estimated reference rankings using bipartite matching based on the Kendall's tau distance. 
\vspace*{-1ex}
\begin{figure}[!htb]
\centering
\includegraphics[width=0.90\linewidth]{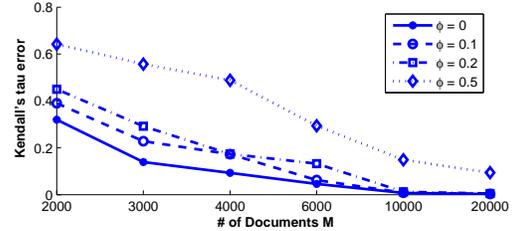}
\vglue -2ex
\caption{{\small The normalized Kendall's tau distance of the
    estimated reference rankings, as functions of $M$, from the semi-synthetic dataset with $Q=100, N=300, K=10$ and different $\phi$.}}
\label{fig:synthetic}
\vspace*{-2ex}
\end{figure}
\vglue -1ex

The ground-truth reference rankings are obtained from a real world movie rating dataset, Movielens, using the same approach as in \cite{topicRank2:ref} over $Q=100$ items and $K=10$. 
We set the same dispersion parameter for all Mallows components as $\phi_k = \phi$ for $\phi = 0, 0.1, 0.2, 0.5$. We use symmetric Dirichlet prior with concentration $\alpha_0 = 0.1$ to generate ranking weights $\bm{\theta}_m$'s. $N =300$. $\mu_{i,j} = 1/{Q\choose 2}, \forall i,j$. The error is further normalized by $W=Q(Q-1)$ and averaged across the $K$ reference rankings. 

Fig.~\ref{fig:synthetic} depicts how the estimation error varies with the number of users $M$ with different values of dispersion. We can see that the reconstruction error in reference rankings for $\phi=0,0.1,0.2$ converges to zero at different rates as a function of $M$. For M4 with $\phi = 0.5$, it converges to a small but non-zero number when $M \rightarrow \infty$. We note that for the ground-truth ranking matrix $\bm{\beta}$, it is $\lambda = 0, 0.01,0.05,0.20$ approximate separable for $\phi=0,0.1,0.2,0.5$ respectively. Our approach therefore can correctly detect the reference rankings when $\lambda$ is small. When $\lambda$ is mild, it can still detect most of the reference rankings correctly.  \footnote{For a random $\bm{\beta}$ with $Q=100,K=10$, it is $0.05$-approximate separable with probability $.933, .870, .793, .426$ for $ \phi = 0,0.1,0.2,0.5$ in a 1000 Monte Carlo runs.}
\vspace*{-2ex}
\subsection{Comparison prediction - Movielens}
\vspace*{-1ex}
\label{sec:predictionlikelihood}
We consider in this section prediction of pairwise comparisons in a benchmark real-world dataset, Movielens.\footnote{ \url{http://grouplens.org/datasets/movielens/}} The star rating dataset is selected due to public availability and widespread use, but we convert it to pairwise comparisons and focus on modeling from the partial ranking viewpoint, as suggested in the ranking literature \cite{Lu11:ref, volkovs14a:ref, topicRank2:ref}. 

We focus on the $Q=200$ most frequently rated movies in the Movielens,  split the first $M=4000$ users for training, and use the remaining users for testing \citep{Lu11:ref}. We convert the training and test ratings into comparisons independently: for all pairs of movies $i,j$ user $m$ rating, $w_{m,n}= (i,j)$ is added if the star ratings for $i$ is higher than $j$, and all ties are ignored. The prior is set to be Dirichlet and it is estimated using methods in \cite{Arora2:ref} given estimated $\widehat{\bm{\beta}}$. 

\begin{figure}[!htb]
\centering
\includegraphics[width=0.85\linewidth]{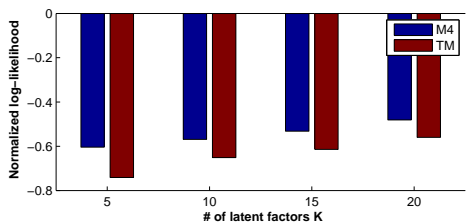}
\vspace*{-1ex}
\caption{{\small The normalized predictive log-likelihood for various $K$ on the truncated Movielens dataset.}}
\label{fig:new-user}
\vspace*{-2ex}
\end{figure}

We evaluate the performance by the {\bf held-out log-likelihood}, i.e., $\Pr(\mathbf{w}_{test}\vert\widehat{\bm{\beta}})$. The log-likelihoods are calculated using the standard Gibbs Sampling approximation in \citep{Wallach09:ref}. The log-likelihoods are then normalized by the total number of comparisons in the testing phase. We compared our new model (M4) against the topic modeling based model in \cite{topicRank2:ref} (TM) with closest settings to our model. We summarize the predictive probability for different $K$ in Fig.~\ref{fig:new-user}. One can see that M4 improves the prediction accuracy of TM for different choice of $K$ and can better fit the real-world observations. 
\vspace*{-1ex} 
\subsection{ Rating prediction via ranking model - Movielens}
\label{sec:rmse}
\vspace*{-1ex}
\begin{table}[!tbh]
\vspace*{-2ex}
\caption{\small Testing RMSE on the Movielens dataset}
\label{table:rmse}
\centering{
\begin{tabular}{|c|c|c|c|c|c|}
\hline 
$K$ & PMF & BPMF &BPMF-int &  TM  & M4 \\ 
\hline 
10 & 1.0491 & 0.8254 &  0.8723 &0.8840 & 0.8509 \\ 
\hline 
15 & 0.9127 & 0.8236 &  0.8734 &0.8780 & 0.8296\\ 
\hline 
20 & 0.9250 & 0.8213 &  0.8678 &0.8721 & 0.8241 \\ 
\hline 
\end{tabular}
 }
\vglue -1ex
\end{table}
To further demonstrate that our model can capture real-world user behavior, we consider the standard rating prediction task in recommendation system \citep{Netflix:ref}. We first train M4 using the training comparisons, and then predict ratings by aggregating the prediction of properly defined test comparisons.  The purpose of this experiment is not to optimize to achieve the best empirical result in the rich literature on rating prediction. 
We use the same training/testing rating split from \citep{BPMF:ref}, and focus on the $Q=100$ most rated movies in Movielens following \cite{topicRank2:ref}. 
We convert the training ratings into training comparisons (for each user, all pairs of movies she rated in the training set are converted into comparisons based on the stars and the ties are ignored) and train a M4 model. The ranking prior is set to be Dirichlet.
To predict stars rating $r_{i,m}$ of user $m$ for movie $i$, we consider the following method: for $s=1,\ldots, 5$, we set $r_{i,m} = s$, and compare it against the movies user $m$ has rated in the training set.  This generates a set of pairwise comparisons $\mathbf{w}_{i,m}(s)$.
For example, if user $m$ has rated movies $A,B,C$ with $4,2,5$ stars respectively in the training set and we are predicting her rating for movie $D$. Then for $s=3$, $\mathbf{w}_{D,m}(3) = \{ (A,D), (D,B), (C,D) \}$. We choose $s$ such that,
\vspace*{-1ex}
\begin{equation*}
\hat{r}_{i,m} = \arg\max_{s} p(\mathbf{w}_{i,m}(s) \vert \mathbf{w}_{train},\widehat{\bm{\beta}}).
\vspace*{-1ex}
\end{equation*}
\vglue -2ex

%
We evaluate the performance using the standard root-mean-square-error ({\bf RMSE}) metric \citep{Netflix:ref}. We compared our approach, M4, against the topic modeling based methods in \cite{topicRank2:ref} (TM), and two benchmark rating-based algorithms, Probability Matrix Factorization (PMF) in \citep{PMF:ref}, and Bayesian probability matrix factorization (BPMF) in \citep{BPMF:ref} that have robust empirical performance 
\footnote{We use the suggested settings to optimize the hyper-parameters and use the implementation and data split from \url{http://www.cs.toronto.edu/~rsalakhu/BPMF.html}}.
Both PMF and BPMF are latent factor models and the number of latent factors $K$ has the similar interpretation as in M4. Note that the ratings predicted by our algorithm are integers from $1$ to $5$, we also round the output of BPMF to the nearest integers from $1$ to $5$ (BPMF-int).

We report the RMSE for different choices of $K$ in Table~\ref{table:rmse}. It is clear that M4 improves upon the ranking-based TM in which the latent factors are restricted to single permutations. On the other hand, when compared to the rating based algorithms, the RMSE of our M4 approach can match BPMF and outperforms BPMF-int and PMF although they are coming from a different feature space. We note that the BPMF typically provides robust and benchmark results on real-world problems. This demonstrates that our approach can accommodate noisy real-world user behavior.

\bibliographystyle{icml2015}
\setlength{\bibsep}{1pt}
\bibliography{ref}
\appendix
\section{Proof for Proposition 1 in the main paper}
We first consider the property of the ranking matrix $\bm{\beta}$ for M4 as summarized in Proposition 1 in the main paper. 
Recall that the ranking matrix $\bm{\beta}$ in M4 is defined as 
\begin{align}
\label{eqa:rankingmatrix}
\beta_{(i,j),k} := \sum_{\sigma:~\sigma(i)<\sigma(j)} p_{\text{M}}(\sigma \vert \sigma_{k}, \phi_k)
\end{align}
%

{\bf Proposition 1 (in the main paper)}
Let the ranking matrix $\bm{\beta}$ be defined as in Eq.~\eqref{eqa:rankingmatrix}, and $\sigma_{k},\phi_k$'s are parameters of the $K$ Mallows distribution. Then, $\forall (i,j)$ and $\forall k$, we have,
\vspace*{-4ex}
\begin{enumerate}
\item[a.] $\beta_{(i,j),k} = \frac{B_{(i,j),k}}{B_{(i,j),k} + B_{(j,i),k}}$
\vspace*{-1ex}
\item[b.] If $\sigma_k(i) < \sigma_k(j)$ and $\phi_k < 1$, $\beta_{(i,j),k} > 0.5 > \beta_{(j,i),k}$
\item[c.] If $\sigma_k(j) = \sigma_k(i) + 1$ and $\phi_k < 1$, $1/\beta_{(i,j),k}  = 1+\phi_k$
\end{enumerate}
\begin{proof}
For $a)$, 
\begin{align*}
\frac{B_{(i,j),k}}{B_{(i,j),k} + B_{(j,i),k}} = \frac{\mu_{i,j}\beta_{(i,j),k}}{\mu_{i,j}\beta_{(i,j),k} + \mu_{j,i}\beta_{(j,i),k}} = \beta_{(i,j),k}
\end{align*}
since $\mu_{i,j} = \mu_{j,i}$ and $\beta_{(i,j),k} + \beta_{(j,i),k} = 1$. 
The proof of $b),c)$ can be derived from the proof for Proposition~3 in the main paper. (see next section)
\end{proof}
\section{Proof for Proposition 3 and Lemma 3 in the main paper}
We first proof the Proposition 3 in the main paper.

{\bf Proposition 3 (in the main paper)}
Let $\sigma_k(i)$ and $\sigma_k(j)$ be the positions of items $i$ and $j$ in the reference ranking $\sigma_k$ of the $k$-th Mallows component. $\phi_k < 1$. If $\sigma_k(i) < \sigma_k(j)$ and $L = \sigma_k(i)-\sigma_k(j) + 1$, then,
\begin{equation}
\frac{\beta_{(j,i),k}}{\beta_{(i,j),k}} \leq L\phi_k^{L-1}
\end{equation}
\begin{proof}
Due to the symmetry in the ranking space, we consider  $\sigma_k(i)=i$ hence $\sigma_k : 1\succ 2\succ \cdots \succ Q$ where $\succ$ indicates ``prefer over''. Instead of directly calculate summation as in the definition,
\vspace*{-1ex}
\begin{align*}
\sigma_{(i,j),k} = \sum_{\sigma:~\sigma(i)<\sigma(j)} p(\sigma \vert \sigma_{k}, \phi_k) 
\end{align*}
\vglue -2ex
we consider the {\it Repeated Insertion Model} (RIM) procedure. RIM is a generative procedure for sampling a ranking which is equivalent to sampling a ranking from a Mallows component. Specifically, in RIM, a ranking $\sigma$ is obtained by sequentially placing the $i$-th item in the reference permutation ($\sigma_k$) into the $j_i$-th position (of the current partial sequence of length $i$), $1\leq j_i\leq i$, in a probabilistic fashion:
\begin{align*}
p_{i}(j_i = l)= \frac{\phi^{i-l}}{1+\phi^{l}+\ldots + \phi^{i-1}}
\end{align*}
and $l\leq i$, $1\leq i\leq Q$.

Let $i<j$. By definition $\beta_{(i,j),k}$ is the probability that item $j$ is inserted after item $i$ in the RIM procedure. 
According to the procedure of RIM, this probability is irrelevant to the items after $j$ and by symmetric, it is irrelevant to the items before $i$. Without loss of generality,  we set $i=1$ and consider $1 < j\leq Q$. For simplicity, we denote $\phi_k = \phi < 1$. 
We first consider $q_{r,s}$, the probability of item $1$ being on the $r$-th position in the sequence after inserting the $s$-th item. $1\leq s\leq j$ and $1\leq r \leq s$. By induction, we shall show that $q_{r,s}=\frac{\phi^{r-1}}{1+\phi^{1} + \cdots + \phi^{s-1}}$. 

As a initial point, after inserting the second item when $s=2$, $q_{1,s} = \frac{1}{1+\phi}$ and $q_{2,s} =\frac{\phi}{1+\phi}$. 
Assume for all $s=1,\ldots, s$, the assumption hold true, then for $s+1$, and $1 < r < s+1$, (i.e., after inserting the item $s+1$)
\vspace*{-1ex}
\begin{align*}
q_{r,s+1} = q_{r,s}\Pr(j_{s+1} > r) + q_{r-1,s}\Pr(j_{s+1} < r)
\end{align*}
where $j_{s+1}$ is the position of item $s+1$ after inserting it into the partial sequence. By the induction assumption, 
\begin{align*}
& q_{r,s}=\frac{\phi^{r-1}}{1+\phi^{1} + \cdots + \phi^{s-1}} \\
& q_{r-1,s}=\frac{\phi^{r-2}}{1+\phi^{1} + \cdots + \phi^{s-1}}
\end{align*}
Therefore,
\begin{eqnarray*}
q_{r,s+1} & =& \frac{\phi^{r-1}}{1+\phi^{1} + \cdots + \phi^{s-1}}\Pr(j_{s+1} > r) \\
& &+ \frac{\phi^{r-2}}{1+\phi^{1} + \cdots + \phi^{s-1}} \Pr(j_{s+1} < r)\\
&= & \frac{\phi^{r-1}}{1+\phi^{1} + \cdots + \phi^{s-1}}\frac{1+\phi+\cdot + \phi^{s-r-1}}{1+\cdot + \phi^{s}} \\
& &+  \frac{\phi^{r-2}}{1+\phi^{1} + \cdots + \phi^{s-1}} \frac{\phi^{s-r+1} + \cdot +\phi^{s}}{1+\cdot + \phi^{s}} \\ 
& =& \frac{\phi^{r-1}}{1 + \cdots + \phi^{s+1-1}}
\end{eqnarray*}
Similarly, it is true for $r=1$ and $r=s$. This conclude the induction hypothesis that, $q_{r,s}=\frac{\phi^{r-1}}{1+\phi^{1} + \cdots + \phi^{s-1}}$.

Now we can calculate $\beta_{(1,j),k}$,
\begin{eqnarray*}
 \beta_{(1,j),k} &=  &\sum\limits_{r=1}^{j-1} q_{r,j-1}\Pr(j_{j}>r) \\ 
& = & \sum\limits_{r=1}^{j-1} \frac{\phi^{r-1} (1+\cdots + \phi^{j-r-1})}{(1+\cdots + \phi^{j-2})(1+\cdots + \phi^{j-1})} \\
& = & \frac{\sum\limits_{r=1}^{j-1}\sum\limits_{l=r-1}^{n-2}\phi^{l}}{(1+\cdots + \phi^{j-2})(1+\cdots + \phi^{j-1})} \\
&= & \frac{1 - j\phi^{j-1} + (j-1)\phi^{j}}{(1-\phi)^2 (1+\cdots + \phi^{j-2})(1+\cdots + \phi^{j-1})}
\end{eqnarray*}
Similarly, we have, 
\begin{eqnarray*}
\beta_{(j,1),k} = \frac{j-1 - j\phi + \phi^{j}}{\phi^{j-1}(1-\phi)^2 (1+\cdots + \phi^{j-2})(1+\cdots + \phi^{j-1})}
\end{eqnarray*}
Therefore, 
\begin{eqnarray*}
\frac{\beta_{(1,j),k}}{\beta_{(j,1),k}} &=& \frac{1 - j\phi^{j-1} + (j-1)\phi^{j}}{\phi^{j-1}(j-1 - j\phi + \phi^{j})} \geq \frac{1}{j\phi^{j-1}}
\end{eqnarray*}
and this conclude our proof. 

We note that in the above equation, if we set $j=2$, we got  $\frac{\beta_{(1,j),k}}{\beta_{(j,1),k}} = \frac{1}{\phi}$. This proves Proposition 1 c.  We also note that $\frac{\beta_{(1,j),k}}{\beta_{(j,1),k}} > 1$ so $\beta_{(1,j),k} > 0.5 > \beta_{(j,1),k}$. This proves Proposition 1 b. 
\end{proof}

Now, we consider the Lemma~3 in the main paper that shows the inevitability of the approximate separability of a random M4. 

{\bf Lemma 2 (in the main paper)}
Let the $K$ reference rankings $\sigma_1,\ldots, \sigma_K$ be sampled i.i.d uniformly from the set of all permutations, and the dispersion parameters $\phi_k <\phi < 1, k=1,\ldots, K$.
Then, the probability that the ranking matrix $\bm{\beta}$ being $\lambda$-approximately separable is at least 
\vspace*{-1ex}
\begin{align}
\label{eqa:p1inTMM}
1 - K \exp( - \frac{Q}{ L(\phi,\lambda)^{2K -1}} )
\end{align} 
\vglue -1ex
where $L(\phi,\lambda) = \text{ceil}\left( (1 + \frac{\log(\lambda)}{\log(\phi)})(1+\epsilon) \right)$ for some positive constant $\epsilon$, and $\text{ceil}(x)$ is the minimum integer that is no smaller than $x$.  
\begin{proof}
Note that by Proposition 3 in the main paper, if $i$ is preferred over $j$ in $\sigma_1$ and under $j$ in other central permutations and the distance of their positions are $L$, then, the corresponding row is at most $L\phi^{L-1}$ approximately novel row for the first topic. This is same for all the topics. 

We note that if we consider two groups of disjoint items, then, the relative rankings within each group is independent to the other group if the ranking is sampled uniformly from all the permutations. In general, we divide the $Q$ items into $Q/L$ groups of disjoint items, each containing $L$ items, denoted by $\{i_{t,1},\ldots, i_{t,L}\}$, for $t=1,\ldots, Q/L$. If a center permutation $\sigma_k$ is sampled uniformly random from the set of all permutations, then, all the partial rankings within each group $t$ are independent to that of another group $s$. 

We now consider for each of these $L$-tuples, the probability that there exist two items $i,j$ such that $i$ is first and $j$ is last in the group for first central permutation $\sigma_1$, and in the opposite way for the other permutations. We denote this probability by $p_1(\phi;\lambda,k)$. By definition, we have, 
\begin{align*}
p_1(\phi;\lambda,k) \geq & \Pr\{ \exists i,j \in\{i_{t,1},\ldots, i_{t,L} \}, s.t., \sigma_{1}(i)<\ldots<\sigma_{1}(j), \\
&\quad \quad \sigma_{2}(i)>\ldots>\sigma_{2}(j), \ldots, \sigma_{K}(i)>\ldots>\sigma_{K}(j) \}\\
= & L(L-1)\left( \frac{1}{(L(L-1))} \right)^{K} = (L(L-1))^{-(K-1)}
\end{align*}

Now, let $\mathcal{B}_k, k=1,\ldots, K$ denote the event that none of the $Q/L$ groups has a $\lambda$-approximately separable row, then, following the same argument as in Lemma~\ref{lem:ranking1}, we have,  
\begin{align*}
\Pr(\bigcup\mathcal{B}_k)\leq K \exp(-Q p_1/L) \leq K \exp(-\frac{Q}{L^{2K-1}})
\end{align*}
as a upper bound for the probability of $\beta$ note being separable. 
We require $L = L(\phi,\lambda)$ such that $L\phi^{L-1} \leq \lambda$. This concludes the proof.  
\end{proof}
\section{Analysis of Proposed Algorithm 1 in the main paper}
Now we formally prove that if a ranking matrix $\bm{\sigma}$ is $\lambda$-approximately separable where $\lambda$ being small enough, the proposed Algorithm 1 can consistently estimate the reference rankings of the shared Mallows components. 

\noindent{\bf Indexing convention:} For convenience, for the rest of
this appendix we will index the $W=Q(Q-1)$ rows of $\mathbf{B}$ and
$\mathbf{E}$ by just a single index $i$ instead of an ordered pair
$(i,j)$ as in the main paper.

\subsection{Consistency of Algorithm 2}
Recall that $\mathbf{E} = \bar{\mathbf{B}}\mathbf{Y}$. We decouple the effect of $\lambda$-separability from the error in estimating $\mathbf{E}$. Note that the second error converges to 0 as $M,N\rightarrow \infty$, we shall focus on the perturbation on solid angle as a result of the $\lambda$-approximate separability. 

For $i$ being a $\lambda$-approximate novel row, let $\mathbf{E}_{i}^{0} = \mathbf{Y}_k$ as the corresponding row of $\mathbf{Y}$. Otherwise, let $\mathbf{E}_{i}^{0} = \mathbf{E}_{i}$ be the rows of $\mathbf{E}$. For each approximate novel row $i$, define the original solid angle as, 
\begin{align}
\label{eqa:ideal-solid-angle}
q_{i}^{0} = \Pr\left( \forall j: \Vert \mathbf{E}_j^{0} - \mathbf{E}_i^{0} \Vert \geq d~:~  \mathbf{E}_i^{0} \mathbf{u} -\mathbf{E}_j^{0} \mathbf{u} >0 \right)
\end{align}
and define the $\lambda$-approximate solid angle as,
\begin{align}
\label{eqa:gamma-solid-angle}
q_{i} = \Pr\left( \forall j: \Vert \mathbf{E}_j - \mathbf{E}_i \Vert \geq d~:~  \mathbf{E}_i \mathbf{u} -\mathbf{E}_j \mathbf{u} >0 \right)
\end{align}
for $i$ being a $\lambda$ approximately novel row. Therefore, for any constant $c>0$, 
\begin{align}
\label{eqa:decomp-ideal-solidangle}
\nonumber
\vert q_{i}^{0} - q_i \vert \leq & \Pr\left( \exists j, *, \vert \mathbf{E}_i^{0} \mathbf{u} -\mathbf{E}_j^{0} \mathbf{u} -  \mathbf{E}_i \mathbf{u} + \mathbf{E}_j \mathbf{u}\vert  \geq c \right) \\
& + \Pr( \forall j, *, \vert \mathbf{E}_{i}^{0} \mathbf{u} -\mathbf{E}_{j}^{0} \mathbf{u} \vert \leq c )
\end{align}
where we have replace the distance constraints with $*$ for convenience. We note that $\mathbf{E}_{j}^{0} = \sum_{k=1}^{K}\bar{B}_{jk} \mathbf{Y}_k$. Without loss of generality, assume that $i$ is a $\lambda$-approximate novel row for $\mathbf{Y}_1$, then, $\mathbf{E}_{i}^{0} = \mathbf{Y}_1$. Taking a closer look at the second term in the above equation, we have, 
\begin{align*}
\vert \mathbf{E}_{i}^{0} \mathbf{u} -\mathbf{E}_{j}^{0} \mathbf{u} \vert = & \vert \sum_{k=2}^{K} \bar{B}_{jk} (\mathbf{Y}_{k} - \mathbf{Y}_{1} ) \mathbf{u}\vert \\
\leq & \sum_{k=2}^{K}\bar{B}_{jk} \vert (\mathbf{Y}_{k} - \mathbf{Y}_{1} ) \mathbf{u}\vert
\end{align*}
And note that $\mathbf{Y}_k, k =2,\ldots, K$ are among the $\mathbf{E}_{j}^{0}$'s, therefore, the second term in \eqref{eqa:decomp-ideal-solidangle} is equivalent to $\Pr(j=k,\ldots, K, \vert (\mathbf{Y}_{k} - \mathbf{Y}_{1} ) \mathbf{u}\vert \leq c)$ hence by union bounding, we have, 
\begin{align*}
\Pr( \forall j, *, \vert \mathbf{E}_{i}^{0} \mathbf{u} -\mathbf{E}_{j}^{0} \mathbf{u} \vert \leq c ) \leq \sum_{k=2}^{K} \Pr( \vert (\mathbf{Y}_{k} - \mathbf{Y}_{1} ) \mathbf{u}\vert \leq c)
\end{align*}

Note that $(\mathbf{Y}_{k} - \mathbf{Y}_{1} ) \mathbf{u} \sim \mathcal{N}(0, \Vert \mathbf{Y}_{k} - \mathbf{Y}_{1} \Vert_{2}^{2} )$, by the property of Gaussian distribution, we have, 
\begin{align*}
& \Pr  (\vert (\mathbf{Y}_{k} - \mathbf{Y}_{1} ) \mathbf{u}\vert \leq c )\\
 = & \int_{-c}^{c} \frac{1}{\sqrt{2\sigma}\Vert \mathbf{Y}_{k} - \mathbf{Y}_{1} \Vert}e^{-t^2/2\Vert \mathbf{Y}_{k} - \mathbf{Y}_{1} \Vert^2} dt \leq \frac{c}{\Vert \mathbf{Y}_{k} - \mathbf{Y}_{1} \Vert}
\end{align*}
For now we denote by $\rho_{\min}$ the minimum of $\Vert \mathbf{Y}_k - \mathbf{Y}_l \Vert $, therefore, the second term in \eqref{eqa:decomp-ideal-solidangle} can be upper-bound by $\frac{c(K-1)}{\rho_{\min}}$. 

For the first term in \eqref{eqa:decomp-ideal-solidangle}, let $\mathbf{e}_{i,j}=\mathbf{E}_i^{0}  -\mathbf{E}_j^{0}   -  \mathbf{E}_i   + \mathbf{E}_j $ and note that $\mathbf{e}_{i,j}\mathbf{u} \sim \mathcal{N}(0, \Vert \mathbf{e}_{i,j} \Vert_{2}^{2})$, then, 
\begin{align*}
\Pr  ( \vert \mathbf{e}_{i,j}\mathbf{u} \vert \geq c ) = 2Q(c/\Vert \mathbf{e}_{i,j} \Vert)\leq \exp(-c^2/2 \Vert \mathbf{e}_{i,j} \Vert_2^2)
\end{align*} 
Further, $\Vert \mathbf{e}_{i,j} \Vert \leq \Vert \mathbf{E}_i^{0} -  \mathbf{E}_i \Vert + \Vert \mathbf{E}_j^{0} - \mathbf{E}_j \Vert$. For $j$ which is not a $\lambda$-approximate novel row and is one of the $j$'s in \eqref{eqa:gamma-solid-angle}, $\Vert \mathbf{E}_j^{0} - \mathbf{E}_j \Vert = 0$. For $j$ being a $\lambda$-approximate novel row and is one of the $j$'s in \eqref{eqa:gamma-solid-angle}, hence $j$ correspond to another topic. Therefore, by the same argument, 
\begin{align*}
\Vert \mathbf{E}_{i}^{0} - \mathbf{E}_{i} \Vert & =\Vert \mathbf{Y}_1 - \sum_{k=1}^{K}\bar{B}_{ik} \mathbf{Y}_k \Vert  \leq \sum_{k=2}^{M} \bar{B}_{ik} \Vert \mathbf{Y}_{1} - \mathbf{Y}_{k} \Vert \leq \\
& \leq \lambda \sum_{k=2}^{M} \Vert \mathbf{Y}_{1} - \mathbf{Y}_{k} \Vert
\end{align*}
Combining the steps together, for Eq.~\eqref{eqa:decomp-ideal-solidangle}, we require, 
\begin{align*}
\vert q_{i}^{0} - q_{i} \vert \leq \frac{c (K-1)}{\rho_{\min}} + W\exp(-[\frac{c}{\lambda K \rho_{\max}}]^{2}) \leq q_{\wedge}/3
\end{align*}
where $q_{\wedge}$ is the minimum solid angle of $\mathbf{Y}$. This is require so that the estimated solid angle for the $\lambda$-approximate novel rows is well-separated from the solid angle of the remaining non-novel rows. 
Recall that $\rho_{\min}$ and $\rho_{\max}$ is defined as the minimum and maximum values of $\Vert \mathbf{Y}_i - \mathbf{Y}_j \Vert,1\leq i \neq j\leq K$. 
To parse the above equation, we set $c= \frac{q_{\wedge} \rho_{\min}}{3K}$ and therefore, we require
\begin{align*}
\lambda \leq \frac{q_{\wedge}\rho_{\min}}{3K^2 \rho_{\max}\sqrt{\log(W/q_{\wedge})}} \leq \frac{q_{\wedge} \kappa}{3K^2 \sqrt{\log(W/q_{\wedge})}}
\end{align*}

We can now apply the same argument to the other rows $i$ whose $d$-neighbor does not enclose a novel word. We thus require $d \geq 12 \lambda K \sqrt{\log(W/q_{\wedge})}/q_{\wedge}$. To combine the two results, we can set
\begin{align}
\label{eqa:thresholding-setting}
d = \mathcal{O}( \kappa / K)
\end{align}

To summarize the discussion, we have, 
\begin{proposition}
\label{prop:solidangle-gamma-approximation}
If $\lambda$ is small enough such that, 
\begin{align}
\label{eqa:gamma-thresholding}
\lambda \leq \frac{q_{\wedge} \kappa}{3K^2 \sqrt{\log(W/q_{\wedge})}}
\end{align}
with $d$ set as in \eqref{eqa:thresholding-setting}. Then, for $M,N\rightarrow \infty$ and the number of projections $P \rightarrow \infty$, the proposed algorithm can find $\mathcal{O}\left( 2K\sqrt{\log(W/q_{\wedge})}/q_{\wedge}\right) \lambda$-approximately novel rows for $K$ distinct topics. 
\end{proposition}

\subsection{Consistency of Algorithm 3}
We now consider the error accumulated in steps in Algorithm 3 in main paper. Assume the Algorithm 2 is correct, we obtain $K$ row vectors, $\mathbf{E}_j, j=1,\ldots, K$, as $\lambda$-approximate novel pairs for the $K$ distinct Mallows components. Without loss of generality, $\mathbf{E}_j$ approximately novel to the $j$-th Mallows component ($j$-th column).  We further denote by $\mathbf{E}_j^{0}$ the ideal extreme points , i.e., $\mathbf{E}_j^{0} = \mathbf{Y}_j$ for $j=1,\ldots, K$. Note that by definition, 
\begin{align*}
\mathbf{E}_i = \sum\limits_{k=1}^{K} \bar{B}_{ik} \mathbf{E}_k^{0}
\end{align*}  
For $i = 1,\ldots, K$, $k\neq i$, we have ${\bar{B}_{ik}} \leq \lambda\bar{B}_{ii}$. $\bar{\mathbf{B}}$ is a row-stochastic matrix.
For $i=1,\ldots, W$, the corresponding row vector $\bar{\mathbf{B}}_{i}$ is the optimal solution of the following constrained linear regression, 
\begin{align*}
{\mathbf{b}}^{*} =  \arg\min_{b_j \geq 0, \sum b_j =1} \Vert \mathbf{E}_i -\sum\limits_{j=1}^{K} b_j \mathbf{E}_j^{0} \Vert
\end{align*}
Now consider the empirical version we have access to which is,
\begin{align*}
\widehat{\mathbf{b}}^{*} =  \arg\min_{b_j \geq 0, \sum b_j =1} \Vert \widehat{\mathbf{E}}_i -\sum\limits_{j=1}^{K} b_j \widehat{\mathbf{E}}_j\Vert
\end{align*}
To bound the error between $\widehat{\mathbf{b}}^{*} $ and ${\mathbf{b}}^{*}$ due to approximate separability, we can establish the following property:
\begin{proposition}
\label{prop:constraint-least-square}
Suppose that for $j = 1,\ldots, K$, $\Vert \widehat{\mathbf{E}}_j - 	\mathbf{E}_j^{0} \Vert_2 \leq \delta_1 $  and $\Vert \widehat{\mathbf{E}}_i - \mathbf{E}_i \Vert_2 \leq \delta_2 $ a fixed $i$. Assume also that $\widehat{\mathbf{E}}_j, j =1,\ldots, K$ are at most $\lambda$-approximately separable and \hbox{$(K-1) \lambda \leq 1$}, then,
\vspace*{-2ex} 
\begin{align*}
\Vert \widehat{\mathbf{b}}^{*} - {\mathbf{b}}^{*} \Vert_2 \leq 4 \frac{\delta_1+\delta_2}{(1- (K-1)\lambda) \lambda_{\min}}
\end{align*}
\vglue -2ex
where $\lambda_{\min}$ denotes the minimum eigenvalue of $\bar{\mathbf{R}}$.
\end{proposition}
\vspace*{-2ex}
\begin{proof}
Let $f(\mathbf{E}^{0}, \mathbf{b}) = \Vert \mathbf{E}_i -\sum\limits_{j=1}^{K} b_j \mathbf{E}_j^{0} \Vert $ for any $\mathbf{b}$ and note that for the optimal solution $\mathbf{b}^{*}$, $f(\mathbf{E}^{0}, \mathbf{b}^{*}) = 0$. Let $\mathbf{Y} = \left[ \mathbf{E}_{1}^{0\top},\ldots, \mathbf{E}_{K}^{0\top}  \right]^{\top}$, we have, 
\begin{align*}
& f(\mathbf{E},\mathbf{b}) - f(\mathbf{E},\mathbf{b}^{*})  =\Vert {\mathbf{E}}_i -\sum\limits_{j=1}^{K} b_j {\mathbf{E}}_j^{0} \Vert -0 \\
=& \Vert \sum\limits_{j=1}^{K} (b_j - b_j^{*}) {\mathbf{E}}_j^{0} \Vert =\sqrt{ (\mathbf{b} - \mathbf{b}^{*}) \mathbf{Y Y^{\top}} (\mathbf{b} - \mathbf{b}^{*})^{\top} }\\
\geq & \Vert \mathbf{b} - \mathbf{b}^{*} \Vert\lambda_{\min, Y}
\end{align*}
Recall that $\mathbf{Y} = \bar{\mathbf{R}}\bar{\mathbf{B}}^{\top}$ and let $\bar{\mathbf{B}}^{\top} = [B_K, B_{r}]^{\top}$ where the $K\times K$ $B_K$ are approximately separable. Note that $B_{K,(i,j)}/B_{K,(i,i)}\leq \lambda$ and $\lambda (K-1)\leq 1$, then, by the Gershgorin circle theorem, the minimum eigenvalue of $B_K$ is lower-bounded by $\frac{1 - (K-1)\lambda}{1 + (K-1)\lambda} > \frac{1- (K-1)\lambda}{2}$. Therefore, $\lambda_{\min, Y} \geq \lambda_{\min} \frac{1- (K-1)\lambda}{2}$ where $\lambda_{\min}$ is the minimum eigenvalue of $\bar{\mathbf{R}}$.
Next, note that for any probability vector $\mathbf{b}$,
\begin{align*}
 \vert f(\mathbf{E},\mathbf{b}) - f(\widehat{\mathbf{E}},\mathbf{b}) \vert \leq & \Vert \mathbf{E}_i -\widehat{\mathbf{E}}_i + \sum b_j (\widehat{\mathbf{E}}_j - \mathbf{E}_j^{0}) \Vert \\
\leq & \Vert \mathbf{E}_i -\widehat{\mathbf{E}}_i \Vert + \sum b_j \Vert \widehat{\mathbf{E}}_j - \mathbf{E}_j^{0} \Vert \\
\leq & \delta_2 + \delta_1
\end{align*}
Combining the above inequalities, we obtain,  
\begin{align*}
\Vert \widehat{\mathbf{b}}^{*} - \mathbf{b}^{*} \Vert \leq & \frac{1}{\lambda_{\min, Y}} \lbrace f(\mathbf{E},\widehat{\mathbf{b}}^{*}) - f(\mathbf{E},{\mathbf{b}}^{*}) \rbrace \\
= &   \frac{1}{\lambda_{\min}, Y} \lbrace f(\mathbf{E},\widehat{\mathbf{b}}^{*}) -f(\widehat{\mathbf{E}},\widehat{\mathbf{b}}^{*}) + f(\widehat{\mathbf{E}},\widehat{\mathbf{b}}^{*}) \\
& ~~ - f(\widehat{\mathbf{E}},{\mathbf{b}}^{*}) +f(\widehat{\mathbf{E}},{\mathbf{b}}^{*})  - f(\mathbf{E},{\mathbf{b}}^{*}) \rbrace \\
\leq &  \frac{1}{\lambda_{\min}, Y} \lbrace f(\mathbf{E},\widehat{\mathbf{b}}^{*}) -f(\widehat{\mathbf{E}},\widehat{\mathbf{b}}^{*}) \\
& ~~ +f(\widehat{\mathbf{E}},{\mathbf{b}}^{*})  - f(\mathbf{E},{\mathbf{b}}^{*}) \rbrace \\
\leq & \frac{4 }{\lambda_{\min}(1-\lambda(K-1))}  (\delta_1 + \delta_2)
\end{align*} 
\end{proof}
\subsection{Consistency of Algorithm 4}
We first consider the row normalization step in Algorithm 3. Note that, $b^{*}(i,j)_k = \bar{B}_{(i,j),k} = \frac{\mu_{i,j}\beta_{(i,j),k}a_k}{\sum \mu_{i,j}\beta_{(i,j),l}a_l}$. We define the row-scaling factor,
\begin{align*}
p_{i,j} = \sum_m X_{(i,j),m} / (\sum_m X_{(i,j),m} + \sum_m X_{(j,i),m})
\end{align*}
and by definition 
$
p_{i,j} \rightarrow \sum \beta_{(i,j),l}a_{l} \leq 1
$
as $M\rightarrow \infty$. If we define $c_{(i,j), k } \leftarrow p_{i,j} b^{*}(i,j)_k$ as intermediate step, and then compute $c_{(i,j),k} / (c_{(i,j),k} + c_{(j,i),k})$. Note that  $c_{(i,j),k} = \beta_{(i,j),k}a_k$ in the ideal case, in order to learn the hidden ranking correctly, we only need $c_{(i,j),k} / (c_{(i,j),k} + c_{(j,i),k}) = \beta_{(i,j),k}$ to remain in the correct interval of either $[0, 0.5]$ or $[0.5,1]$. Therefore, the error in estimating $c_{(i,j),k}$ should satisfy,
\begin{align*}
\vert c_{(i,j),k} - \hat{c}_{(i,j),k} \vert \leq a_k \vert 0.5 - \beta_{(i,j),k} \vert
\end{align*} 
Recall that $p_{i,j}$ can be estimated much accurate than $b^{*}$, Therefore, we can consider the error in $c$ as the result of error in $\hat{b}^{*}$. 
Note that the minimum of the $ \vert 0.5 - \beta_{(i,j),k} \vert$ is achieved if the position of item $i,j$ in the reference ranking are next to each other and $\vert 0.5 - \sigma_{(i,j),k} \vert\geq \frac{1-\phi}{2(1+\phi)} \geq (1-\phi)/4$. Therefore, we require,
\begin{align*}
\vert \hat{b}^{*}(i,j)_k - b^{*}(i,j)_k \vert p_{i,j} \leq a_k (1-\phi)/4
\end{align*}
Let $a_{\min} = \min a_k$ and note that $p_{i,j}<1$, using result in Prop.~\ref{prop:constraint-least-square}, we require,
\begin{align}
\label{eqa:requirement-by-all-rest-steps}
 \delta_1 + \delta_2 \leq a_{\min} \lambda_{\min} (1- (K-1)\lambda) (1-\phi)/8
\end{align}

Now, we express $\delta_1$ and $\delta_2$ in terms of $\lambda$. Note that $\delta_2 = \Vert \widehat{\mathbf{E}}_i - \mathbf{E}_i \Vert$ and $\delta_1 = \Vert \widehat{\mathbf{E}}_j -\mathbf{E}_j^{0} \Vert \leq \Vert \widehat{\mathbf{E}}_j - \mathbf{E}_j \Vert + \Vert \mathbf{E}_j^{0} - \mathbf{E}_j \Vert $. $\delta_2$ and the first term in $\delta_1$ converges to 0 exponentially in $M,N$ and does not depend on $\lambda$. Hence we focus on the term $\Vert \mathbf{E}_j^{0} - \mathbf{E}_j \Vert$. 

Note that $ \Vert \mathbf{E}_j^{0} - \mathbf{E}_j \Vert = \Vert \sum_{k\neq j}\bar{B}_{jk} (\mathbf{E}_k^{0}) - (1-\bar{B}_{jj})\mathbf{E}_j^{0} \Vert $. Let $v = [-(1-\bar{B}_{11}), \bar{B}_{12},\ldots, \bar{B}_{1K}]$ (wlog, consider $j=1$), then, $\Vert \mathbf{E}_j^{0} - \mathbf{E}_j \Vert  \leq \Vert v \Vert \lambda_{\max, Y}$. Following the same steps in Prop.~\ref{prop:constraint-least-square} and denoting $\lambda_{\max}$ to be the maximum eigenvalue of $\bar{\mathbf{R}}$, we have, $\lambda_{\max, Y} \leq (1+ (K-1)\lambda)\lambda_{\max}$, and $\Vert v \Vert \leq \lambda (K-1)/ (1+ (K-1)\lambda) $. Combining the results, we have, 
\begin{align*}
\Vert \mathbf{E}_j^{0} - \mathbf{E}_j \Vert \leq \lambda (K-1)\lambda_{\max}
\end{align*}
Let's consider $K\lambda\ll 1$ and using all the results above, we need,
\begin{align*}
\lambda \leq \frac{a_{min}\lambda_{\min} (1-\phi)}{8K\lambda_{\max}}
\end{align*}
Formally, to combine the above two sections for Algorithm 3 and 4, we have,
\begin{proposition}
\label{prop:regression-scaling-post-results}
Assume $K$ rows that $\lambda$-approximately novel pairs for $K$ distinct Mallows components are selected. The remaining steps, i.e., constrained linear regression, row-scaling, and post-processing can recover the true reference rankings of all Mallows component when $M\rightarrow \infty$ and
\begin{align*}
\lambda \leq \frac{a_{min}\kappa (1-\phi)}{8(K-1)}
\end{align*}
where $a_{\min} = \min_{k}a_k$, $\kappa = \lambda_{\min}/\lambda_{\max}>0$ is the condition number of $\bar{\mathbf{R}}$, and $\phi =\max_{k} \phi_k < 1$. 
\end{proposition}

\subsection{Overall sample complexity of the Algorithm 1 via random projection}
We can directly combine the results from Prop.~\ref{prop:solidangle-gamma-approximation}, \ref{prop:constraint-least-square} and \ref{prop:regression-scaling-post-results} to obtain the consistency results for the overall algorithm. 

{\bf Theorem 2 in the main paper} 
Let the ranking matrix $\bm{\beta}$ be $\lambda$-approximate separable and the second order moments $\mathbf{R}$ of ranking prior to be full rank. If
\begin{align}
\label{eqa:lambda_min}
\lambda \leq  \frac{a_{\min} \kappa (1-\phi) q_{\wedge}}{8 K^{2} a_0 \sqrt{\log(W/q_{\wedge})}}
\end{align}
and $M,P \rightarrow \infty$, then, Algorithm~1 can consistently recover all the reference rankings of the latent Mallows distributions. Moreover, $\forall \delta>0$, if 
\begin{align*}
M \geq \max \Biggl\{\frac{640 W^2 \log(3W/\delta)}{ N \eta^4 d^2 q_{\wedge}^{2}}, ~
\frac{320 W \log(3W/\delta)}{N \eta^4 \lambda_{\min}^{2} a_{\min}^{2} (1-\phi)^{2}} \Biggr\}
\end{align*}
and for 
\begin{equation*}
P \geq 32 \frac{\log(3W/\delta)}{q_{\wedge}^2}
\end{equation*}
the proposed algorithm fails with probability at most $\delta$. 
The other model parameters are defined as follows: $\eta = \min_{1\leq w\leq W} [\mathbf{B}\mathbf{a}]_w$; $a_{\max}$, $a_{\min}$ are the max/min of entries of $\mathbf{a}$; $a_0 =\max_{i,j} a_{i}/a_{j}$; $\mathbf{Y} = \mathbf{R}\bar{\mathbf{B}}$; $\kappa = \lambda_{\min} / \lambda_{\max}$ is the condition number of $\bar{\mathbf{R}}$; $q_{\wedge}$ be the minimum normalized solid angle formed by row vectors of $\mathbf{Y}$; $d = 6\kappa / K$; $\phi_k \leq \phi <1$. $N$ is the number of comparisons of each user.
\begin{proof}
First note that $\bar{B}_{i,k} =\mu_{i}\beta_{i,k}a_k$. Therefore, if $\bm{\beta}$ is $\lambda$-approximately separable, then, $\bar{\mathbf{B}}$ is at most $a_0\lambda$-approximately separable. 

Now, assuming that $\lambda a_0 \leq \frac{q_{\wedge} \kappa}{3K^2 \sqrt{\log(W/q_{\wedge})}} $, by proposition \ref{prop:solidangle-gamma-approximation}, the novel word step via random projection can select roughly $c_1 K\lambda a_0/q_{\wedge}$-approximately separable novel words if $M,N\rightarrow \infty$ and $P \rightarrow \infty$. 

Now apply proposition \ref{prop:regression-scaling-post-results}, we require $c_1 K\lambda a_0/q_{\wedge} \leq \frac{a_{\min} \kappa (1-\phi)}{8K}$, therefore, 
\begin{align*}
\lambda \leq \frac{a_{\min} \kappa (1-\phi) q_{\wedge}}{8 c_1 K^{2} a_0} = \frac{a_{\min} \kappa (1-\phi) q_{\wedge}}{8 K^{2} a_0 \sqrt{\log(W/q_{\wedge})}}
\end{align*}
Note that this is stronger than previous constraints. In sum, given these constraints, and let $M,P\rightarrow \infty$, the estimation on the center rankings are consistent.  

The sample complexity follows directly from results as in \citep{topicRank:ref} except for the constants.
\end{proof}

\end{document}